\documentclass[11pt]{article}

\usepackage[preprint]{acl}

\usepackage{times}
\usepackage{latexsym}
\usepackage{algorithm}
\usepackage{algpseudocode}
\usepackage{pifont}
\usepackage{xcolor}
\usepackage{times}
\usepackage{latexsym}

\usepackage[T1]{fontenc}

\usepackage[utf8]{inputenc}

\usepackage{microtype}

\usepackage{inconsolata}

\usepackage{graphicx}
\usepackage{overpic}
\usepackage{xspace}
\usepackage{enumitem}
\usepackage{multirow}
\usepackage{mdframed}
\usepackage{amsmath}
\usepackage{url}            
\usepackage{booktabs}       
\usepackage{amsfonts}       
\usepackage{nicefrac}       
\usepackage{xcolor}         
\usepackage{nameref}
\usepackage{soul}
\usepackage{color}
\usepackage{colortbl}
\usepackage{algorithm}
\usepackage{algpseudocode}
\usepackage{listings}
\usepackage{tcolorbox} 

\definecolor{darkBlue}{rgb}{0.000000,0.000000,0.545098}
\definecolor{darkGreen}{rgb}{0.000000,0.392157,0.000000}
\definecolor{DarkGray}{gray}{0.4}
\definecolor{javared}{rgb}{0.6,0,0} 
\definecolor{javagreen}{rgb}{0.25,0.5,0.35} 
\definecolor{javapurple}{rgb}{0.5,0,0.35} 
\definecolor{javadocblue}{rgb}{0.25,0.35,0.75} 
\definecolor{lightgray}{gray}{0.7}
\definecolor{lightblue}{rgb}{0.63, 0.79, 0.95}

\definecolor{shadecolor}{RGB}{150,150,150}
\definecolor{blueA}{RGB}{204,229,255}
\definecolor{redA}{RGB}{112,0, 0}
\definecolor{RED}{RGB}{255,0, 0}
\definecolor{lightred}{HTML}{FFEBEE}

\newfloat{example}{htbp}{loe}
\floatname{example}{Example}
\floatstyle{plain} 
\restylefloat{example}

\lstdefinestyle{mystyle}{
  frame=single,
  xleftmargin=4pt,
  xrightmargin=4pt,
  abovecaptionskip=2pt,
  belowcaptionskip=0pt,
  captionpos=b,
  escapeinside={*‘}{’*},
  tabsize=4,
  emphstyle={\bf},
  basicstyle=\linespread{0.9}\small\ttfamily,
  keywordstyle=\color{javapurple}\bfseries,
  stringstyle=\color{javared},
  commentstyle=\color{javagreen},
  morecomment=[s][\color{javadocblue}]{/**}{*/},
  showspaces=false,
  columns=flexible,
  showstringspaces=false,
  morecomment=[l]{//},
  breaklines=true
}

\lstdefinestyle{tblstyle}{
  frame=none,
  xleftmargin=-10pt,
  xrightmargin=0pt,
  abovecaptionskip=0pt,
  belowcaptionskip=0pt,
  captionpos=b,
  escapeinside={*‘}{’*},
  tabsize=4,
  emphstyle={\bf},
  basicstyle=\linespread{0.5}\fontsize{6.5}{8}\ttfamily,
  keywordstyle=\color{javapurple}\bfseries,
  stringstyle=\color{javared},
  commentstyle=\color{javagreen},
  morecomment=[s][\color{javadocblue}]{/**}{*/},
  showspaces=false,
  columns=flexible,
  showstringspaces=false,
  morecomment=[l]{//},
  breaklines=true
}

\definecolor{lightgray}{gray}{0.7}
\definecolor{lightred}{HTML}{FFEBEE}
\definecolor{royalblue}{RGB}{65,105,225} 
\usepackage{xspace}
\makeatletter
\DeclareRobustCommand\onedot{\futurelet\@let@token\@onedot}
\def\@onedot{\ifx\@let@token.\else.\null\fi\xspace}

\makeatother
\definecolor{JungleGreen}{RGB}{78,138,62}
\definecolor{blue}{RGB}{0,102,204}  

\definecolor{lightgray}{HTML}{eeeeee}
\definecolor{darkgray}{HTML}{d8d8d8}
\definecolor{tablecellgreen}{HTML}{d0e6d0}

\definecolor{highlightColor}{rgb}{1, 0.8, 0.6}

\definecolor{amii_magenta}{HTML}{bf477c}
\definecolor{amii_summer}{HTML}{ffcccc}
\definecolor{amii_mustard}{HTML}{faa53c}
\definecolor{amii_sky}{HTML}{6c98ab}
\definecolor{amii_emerald}{HTML}{006c65}
\definecolor{amii_night}{HTML}{003f58}

\definecolor{top1Color}{HTML}{ff770f}
\definecolor{top2Color}{HTML}{01847f}
\definecolor{top3Color}{HTML}{ffecb3}
\definecolor{propColor}{HTML}{002FA7}
\sethlcolor{highlightColor}

\usepackage{subcaption}

\newif\ifdisplaycontent
\displaycontenttrue  

\def\ourmethod{\textsc{LogiSafetyGen}}
\def\ourbench{\textsc{LogiSafetyBench}}

\definecolor{color1}{HTML}{E5E5E1}
\definecolor{color2}{HTML}{E0DDEF}
\definecolor{color3}{HTML}{F3D7CA}
\definecolor{color4}{HTML}{E7E2B6}
\definecolor{color5}{HTML}{EAD6FA}

\newcommand{\circled}[1]{{\large \textcircled{\footnotesize #1}}}

\newcounter{finding}

\usepackage{tcolorbox}
\tcbuselibrary{skins, breakable}

\newtcolorbox{promptbox}[1]{
    colback=blue!3!white,       
    colframe=blue!30!black,     
    coltitle=white,             
    fonttitle=\bfseries\large,  
    title={#1},                 
    sharp corners,              
    boxrule=0.8mm,              
    breakable,                  
    enhanced,                   
    left=5pt, right=5pt, top=5pt, bottom=5pt 
}

\newtcolorbox{agentprofile}[2][]{
    colback=white,              
    colframe=gray!20!black,     
    coltitle=white,             
    fonttitle=\bfseries\sffamily,
    title={#2},                 
    enhanced,
    attach boxed title to top left={yshift=-2mm, xshift=2mm}, 
    boxed title style={colback=black, sharp corners},
    boxrule=0.5mm,
    sharp corners,
    fontupper=\small,           
    #1                          
}

\newcommand\blfootnote[1]{%
  \begingroup
  \renewcommand\thefootnote{}\footnote{#1}%
  \addtocounter{footnote}{-1}%
  \endgroup
}

\title{Evaluating Implicit Regulatory Compliance in LLM Tool Invocation via Logic-Guided Synthesis}

\author{
  \textbf{Da Song}\textsuperscript{1,2},
  \textbf{Yuheng Huang}\textsuperscript{3}\thanks{~Corresponding Author},
  \textbf{Boqi Chen}\textsuperscript{4},
  \textbf{Tianshuo Cong}\textsuperscript{1,2},
  \\
  \textbf{Randy Goebel}\textsuperscript{5},
  \textbf{Lei Ma}\textsuperscript{3,5},
  \textbf{Foutse Khomh}\textsuperscript{6}
}

\begin{document}
\maketitle

\blfootnote{
    \\
    \textsuperscript{1}School of Cryptologic Science and Engineering, Shandong University, Jinan, Shandong, China \\
    \textsuperscript{2}Shandong Key Laboratory of  Artificial Intelligence Security, Shandong University, Jinan, Shandong, China \\
    \textsuperscript{3}The University of Tokyo, Tokyo, Japan \\
    \textsuperscript{4}McGill University, Montreal, Canada \\
    \textsuperscript{5}University of Alberta, Edmonton, AB, Canada \\
    \textsuperscript{6}Polytechnique Montreal, Montreal, Canada \\
    \textbf{Emails:} \{song\_da, tianshuo.cong\}@sdu.edu.cn,
    yuhenghuang42@g.ecc.u-tokyo.ac.jp,
    boqi.chen@mail.mcgill.ca,
    rgoebel@ualberta.ca,
    ma.lei@acm.org,
    foutse.khomh@polymtl.ca
}

\begin{abstract}
    The integration of large language models (LLMs) into autonomous agents has enabled complex tool use, yet in high-stakes domains, these systems must strictly adhere to regulatory standards beyond simple functional correctness. However, existing benchmarks often overlook implicit regulatory compliance, thus failing to evaluate whether LLMs can autonomously enforce mandatory safety constraints.
    To fill this gap, we introduce {\ourmethod}, a framework that converts unstructured regulations into Linear Temporal Logic oracles and employs logic-guided fuzzing to synthesize valid, safety-critical traces. Building on this framework, we construct {\ourbench}, a benchmark comprising 240 human-verified tasks that require LLMs to generate Python programs that satisfy both functional objectives and latent compliance rules. Evaluations of 13 state-of-the-art (SOTA) LLMs reveal that larger models, despite achieving better functional correctness, frequently prioritize task completion over safety, which results in non-compliant behavior.
\end{abstract}

\section{Introduction}

Recent advances in large language models (LLMs) have enabled the emergence of LLM-based agents that can interpret complex user instructions, invoke external tools, and interact with the physical or digital world to achieve multi-step goals~\cite{qin2024toolllm}. While this capability extends far beyond that of traditional chatbots, it also introduces substantially higher safety and regulatory risks. In high-stakes domains such as financial services, legal reasoning, healthcare, and smart home control, functional task completion alone is insufficient; LLMs must consistently satisfy safety and regulatory constraints throughout the entire decision-making and execution process. Violations may result in severe consequences~\cite{ye2024toolsword, radosevich2025mcp}, including physical harm and regulatory non-compliance.

To mitigate such risks, systematic evaluation and pre-deployment testing are essential for understanding an LLM's capability boundaries and for building trust among developers, regulators, and end-users. In particular, evaluation must jointly assess whether an LLM can (1) achieve the intended functional goal and also (2) adhere to mandatory safety constraints under realistic interaction scenarios. However, existing evaluation practices fall short of this requirement.

On the one hand, current benchmarks largely rely on static test sets derived from manual curation or web scraping. These datasets are expensive to scale~\cite{zhang2024darg}, difficult to validate~\cite{huang2025evaluating}, and prone to data saturation~\cite{chen-etal-2025-benchmarking-large}. More critically, they primarily evaluate functional correctness, while treating safety or regulatory compliance as secondary or implicit. 
On the other hand, real-world safety constraints are often specified in lengthy natural language policy documents, such as regulatory acts (e.g., EU AI Act~\cite{EUAIAct}) or internal compliance manuals, making it prohibitively costly to involve domain experts in mapping relevant rules to possible user instructions or execution traces.

To address these limitations, we propose {\ourmethod}, an automated framework for synthesizing high-quality test cases for safety-critical LLMs. {\ourmethod} takes tool specifications and regulation documents as input and automatically generates test scenarios with explicit, verifiable safety constraints. The framework consists of three stages. 
(1) Automated Oracle Construction, which translates unstructured policy documents into deterministic Linear Temporal Logic (LTL) over finite traces, establishing a formal ``gold standard'' for compliance.
(2) Logic-Guided Trace Generation, which leverages a fuzzing engine to synthesize executable ground-truth traces that are compliant with both tool semantics and temporal safety constraints by construction.
(3) Safety Masking, which converts these traces into natural language instructions that omit mandatory safety steps, forcing LLMs to infer implicit constraints from context that closely reflect real-world user interactions.

We instantiate {\ourmethod} in three representative high-stakes domains—Financial Services, Tele-Healthcare, and Smart Home IoT—and construct {\ourbench}, a benchmark of 240 manually verified test cases. Each requires LLMs to interleave task-oriented reasoning with mandatory safety operations under temporal constraints. 

Our contributions are summarized as follows:

\vspace{-0.5em}
\begin{itemize}[leftmargin=*]
\setlength\itemsep{0.3mm}
    \item We introduce {\ourmethod}, an automated benchmark generation framework that integrates LTL-based formal verification with logic-guided fuzzing to produce safety-critical test cases.

    \item We propose a Dual-Oracle Evaluation protocol that jointly verifies functional correctness and strict temporal safety compliance, thus enabling precise measurement of the Inference Gap between task success and rule adherence.

    \item We release {\ourbench}, a challenging, human-verified benchmark spanning three high-stakes domains, which require LLMs to satisfy implicit regulatory constraints during execution.

    \item We conduct a systematic evaluation of SOTA LLMs, showing that although scaling improves general reasoning ability, it does not resolve failures in implicit regulatory compliance.
\end{itemize}
\vspace{-0.5em}

\section{Related Work}

\noindent\textbf{Trustworthiness Evaluation of LLM.} A growing body of work aims to provide benchmarks on the safety, security, and regulatory compliance of LLM-based systems, which is crucial for their trustworthy deployment in real-world~\cite{mohammadi2025evaluation}. Early work, such as AgentHarm~\cite{andriushchenko2025agentharm, 10.5555/3692070.3693501}, analyzes LLM behavior under malicious user requests. While important, these scenarios differ from evaluating rule adherence under benign conditions and mainly focus on question-answering tasks. More recent frameworks, including ToolEmu~\cite{ruan2024identifying}, Agent-SafetyBench~\cite{zhang2024agent}, and AgentBench~\cite{liu2024agentbench}, examine general tool-use safety beyond simple QA. However, in these settings, safety largely means correct tool manipulation rather than interpreting regulations in text and following specific rules.

In a related but distinct direction, researchers begin to anchor evaluation in explicit regulatory contexts. EU-Agent-Bench~\cite{euagentbench2025} and HSE-Bench~\cite{wang2025llm} introduce evaluation scenarios derived from concrete policies and standards, while RuleArena~\cite{zhou2025rulearena} and CNFinBench~\cite{ding2025cnfinbenchbenchmarksafetycompliance} move closer to structured rule reasoning by embedding real-world regulations into their tasks. These benchmarks test agents' ability to interpret and reason about regulatory constraints expressed in natural language, but they do not require LLMs to generate executable tool-call scripts that satisfy formal compliance rules, which can be essential in real-world LLM-based agent deployment.

\begin{figure*}[h!]
    \centering
    \includegraphics[width=0.9\linewidth]{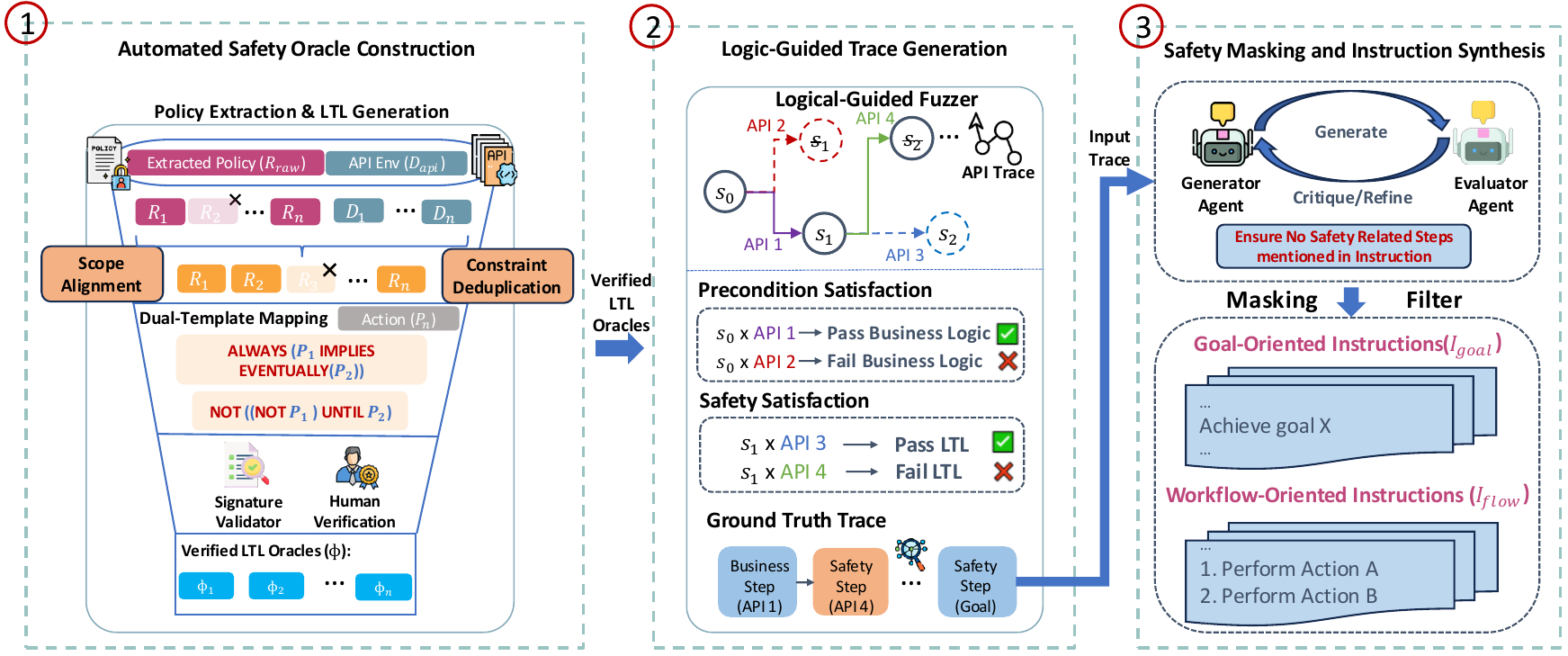}
    \vspace{-0.5em}
    \caption{The {\ourmethod} framework. \circled{1} We translate regulations into grounded LTL oracles. \circled{2} \textit{Logic-Guided Fuzzer} synthesizes compliant traces ($\tau^*$). \circled{3} A safety masking pipeline generates natural language instructions ($I$) that force agents to infer implicit constraints.}
    \label{fig:workflow}
    \vspace{-1.2em}
\end{figure*}

\noindent\textbf{Benchmark Construction Methods.}
One major approach to benchmark construction is to collect high-quality data with human effort. HumanEval~\cite{chen2021codex} is one of the classic examples in this category. Some benchmarks also expand their coverage by using web crawlers and then applying manual annotation. In tool-use evaluation, BFCL~\cite{patil2025the} is a representative example that relies on crawling large-scale repositories. However, since manual benchmarks are both costly and potentially contaminated~\cite{10.5555/3692070.3694661}, researchers have explored LLM-in-the-loop benchmark generation as a more scalable alternative. In these approaches, models help synthesize test instances automatically~\cite{ghazaryan2025syndarin}. Notable examples include AutoCodeBench~\cite{chou2025autocodebench}, which generates code-centric tasks, and MCPBench~\cite{wang2025mcp}, which produces tool-use test cases with LLM assistance. Although these methods reduce annotation effort and broaden evaluation scope, the stochastic nature of LLM generation makes it hard to guarantee strict correctness or compliance with specific rules.

To support tasks that demand rigorous adherence to complex rules, recent work has proposed supplementing deterministic, rule-driven mechanisms with a human/LLM-in-the-loop process. Closest to our setting is ShieldAgent~\cite{chen2025shieldagent}, which uses probabilistic rule circuits to constrain agent trajectories; however, it is designed as a runtime defense against adversarial attacks, whereas our framework serves as an offline stress test of compliance. Similarly, StateEval~\cite{huang2025evaluating} introduces state-aware test generation for sequential API calls, focusing on functional correctness rather than regulatory compliance.

\section{Methodology}
\label{sec:method}

\subsection{Overview}
We formalize the problem of \textit{safety-aware LLM evaluation} as follows: Given a toolset $\mathcal{T}$ and a regulatory policy $\mathcal{P}$, our goal is to automatically synthesize a test case $(I, \mathcal{S}^*, \Phi_{\text{task}})$ where $I$ represents user instructions, $\mathcal{S}^*$ specifies the desired target states, and $\Phi_{\text{task}}$ encodes the applicable regulatory constraints. 
The states $\mathcal{S}^*$ are guaranteed to be reachable by a valid ground-truth API invocation trace $\tau$ which is generated alongside the test case.
To achieve fully automated synthesis, we must overcome three inherent technical challenges:

\noindent\textbf{(1) The Ambiguity Gap.} Regulations in $\mathcal{P}$ are unstructured natural language, and typically lack the required mathematical determinism to support automated verification.

\noindent\textbf{(2) The Validity Gap.} Naively using LLMs to generate traces often results in hallucinated API calls or invalid arguments, rendering the test cases non-executable.

\noindent\textbf{(3) The Inference Gap.} Simply stating safety rules in the instruction $I$ is trivial; a rigorous test must ensure the LLM can \textit{infer} implicit constraints from context, simulating real-world compliance.

We address these three critical gaps via our method {\ourmethod}. As illustrated in Figure~\ref{fig:workflow}, our framework generates test cases by sequentially addressing three critical gaps in compliance testing:
First, \circled{1} Automated Safety Oracle Construction (Section~\ref{sec:regulation-to-logic}) addresses the \textbf{Ambiguity Gap} by translating unstructured regulations into deterministic Linear Temporal Logic ($\text{LTL}_f$) constraints, establishing a verifiable mathematical gold standard ($\Phi$).
Next, \circled{2} the Logic-Guided Trace Generation (Section~\ref{sec:fuzzer}) component bridges the \textbf{Validity Gap} by employing a constraint-satisfaction fuzzing engine to synthesize ground truth traces ($\tau^*$) that are guaranteed to be executable.
Finally, \circled{3} Safety Masking and Instruction Synthesis (Section~\ref{sec:translator}) targets the \textbf{Inference Gap} using a translation-masking strategy to generate user prompts ($I$) that explicitly describe business goals while strictly omitting the safety steps found in $\tau^*$.
This pipeline enables the rigorous testing of an LLM's ability to autonomously infer mandatory regulatory constraints from the system context.

\subsection{Automated Safety Oracle Construction}
\label{sec:regulation-to-logic}
Establishing a gold standard for compliance from massive, unstructured regulatory texts is non-trivial. Unlike robotics approaches restricted to closed vocabularies~\cite{pan2023data}, API-based environments are prone to ambiguity and hallucination. To bridge this \textit{Ambiguity Gap}, we design a pipeline that enforces strict grounding. 

\noindent\textbf{Rule Extraction}. 
We first leverage LLMs to distill regulations into atomic constraints, applying \textit{Scope Alignment} and \textit{Constraint Deduplication}
to retain only relevant policies. Furthermore, rather than allowing unconstrained free-form logical translation with LLMs, we restrict all extracted constraints to two LTL templates that capture the dominant forms of compliance logic in business, as observed in prior work~\cite{chen2025shieldagent}. Specifically, we adopt a dual-template formulation comprising: 
(1) \textbf{\textit{Operational Restrictions}} which forbid a sensitive action $P_2$ until a check $P_1$ holds, formalized as $\neg ((\neg P_1) \mathcal{U} P_2)$;
(2) \textbf{\textit{Instruction Adherence}} requires that a trigger $P_1$ implies a future outcome $P_2$, formalized as $\square (P_1 \rightarrow \diamond P_2)$.

\noindent\textbf{Rule Validation.} 
Template-based generation yields a set of candidate formulas $\Phi_{\text{cand}}$, which may still reference semantically plausible but non-existent predicates (e.g., hallucinated \texttt{verify\_user} vs.\ the concrete \texttt{check\_auth}). To guarantee executability, we subject $\Phi_{\text{cand}}$ to a deterministic \emph{Signature Validator} that eliminates any formula containing predicates not already defined in the API schema. This grounding step ensures that all retained Safety Oracles $\Phi$ are syntactically valid and computationally solvable by the fuzzer, thus completing the automated logic extraction pipeline.

\begin{table*}[t]
\caption{Benchmark Comparison. {\ourbench} uniquely combines logic-guided trace generation, implicit safety testing, and formal LTL oracles within an executable sandbox.}
\vspace{-0.5em}
\centering
\small
\newcommand{\cmark}{\textcolor{green!60!black}{\ding{51}}}%
\newcommand{\xmark}{\textcolor{red}{\ding{55}}}
\resizebox{\textwidth}{!}{
\begin{tabular}{lccccc}
\toprule
\textbf{Benchmark} & \textbf{Primary Focus} & \textbf{Auto. Valid Trace} & \textbf{Implicit Safety} & \textbf{Formal Oracle} & \textbf{Executable Env.} \\
\midrule
ToolSword~\cite{ye2024toolsword} & Adversarial Injection & \xmark & \xmark & \xmark & \cmark \\
TAI3~\cite{feng2025intentest} & Intent Integrity & \cmark & \xmark & \xmark & \cmark \\
ToolEmu~\cite{ruan2024identifying} & General Risks & \xmark & \cmark & \xmark & \xmark \\
Agent-SafetyBench~\cite{zhang2024agent} & General Safety & \xmark & \cmark & \xmark & \cmark \\
\midrule
\rowcolor{gray!10} \textbf{\ourbench (Ours)} & \textbf{Regulatory Compliance} & \textbf{\cmark} & \textbf{\cmark} & \textbf{\cmark} & \textbf{\cmark} \\
\bottomrule
\end{tabular}
}

\label{tab:benchmark_comparison}
\vspace{-1em}
\end{table*}

\subsection{Logic-Guided Trace Generation} \label{sec:fuzzer} 

With the formal Safety Oracles ($\Phi$) established, our next goal is to synthesize concrete execution traces ($\tau$). These ordered tool invocation sequences serve as the ground-truth solutions for our benchmark, providing the ``gold standard'' behavior for how an ideal LLM can interleave functional reasoning with mandatory safety checks to satisfy the constraints in $\Phi$. However, generating the ground-truth sequences creates the \textbf{Validity Gap}. Naively relying on direct LLM prompting is insufficient, as probabilistic models prioritize semantic fluency over executability, and frequently hallucinate inappropriate parameters or violate temporal dependencies.

To bridge this gap, we adopt Fuzzing~\cite{zhu2022fuzzing}, a dynamic testing paradigm that discovers valid execution paths by systematically exploring the state space under strict constraints. Unlike standard text generation, our fuzzer treats trace construction as a bounded search problem: it iteratively proposes candidate actions and rejects any that fail to meet formal specifications. We implement a safety-constrained fuzzer (with $\approx$3600 lines of code) that constructs traces $\tau = [a_1, a_2, ..., a_k]$ incrementally. The core of the fuzzer is a Dual-Constraint Pruning Mechanism. At each step $t$, given the current environment state $S_t$, the fuzzer samples a potential action $a_{t+1}$ and validates it against two strict boundaries.

\noindent\textbf{Precondition Satisfaction ($\mathcal{V}_{\text{schema}}$):} The action must be executable in the current state $S_t$. This ensures that the trace is functionally sound.
    
\noindent\textbf{Safety Satisfaction ($\mathcal{V}_{\text{safe}}$):} The updated trace $\tau_{t+1}$ must satisfy the LTL Safety Oracles. We verify this using a runtime LTL monitor: $\mathcal{V}_{\text{safe}}(\tau_{t+1}) = \text{True} \iff \forall \phi \in \Phi, \tau_{t+1} \models \phi$.

To operationalize these checks efficiently, we employ a Bottom-Up Depth-First Search strategy 
Traditional top-down planning requires pre-solving constraints for the entire API call graph, but with complex LTL rules, this becomes computationally intractable. In contrast, our bottom-up approach enables Lazy Evaluation. We compute the valid search space and check compliance only for the immediate next step ($S_t \to S_{t+1}$). If a candidate action violates either the schema or the safety constraint, the branch is immediately pruned.

When the fuzzer attempts a business action that violates a safety constraint (LTL), the LTL monitor triggers a violation ($\mathcal{V}_{\text{safe}} = \text{False}$), pruning that path. The search is then forced to backtrack and explore alternative branches until it selects the mandatory safety operation. By iterating this process until the target trace length is reached, the fuzzer produces a ground truth trace $\tau^*$ where every user action is wrapped in the necessary safety checks, guaranteeing that the test case is solvable, executable, and strictly compliant by construction. 

\subsection{Safety Masking and Instruction Synthesis}
\label{sec:translator}
Phase~\circled{2} yields a ground truth trace $\tau^*$ that is executable and fully compliant. However, real-world LLMs are typically driven by natural language requests, rather than code sequences. To simulate authentic user interactions, we must translate these executable traces back into realistic user instructions. Critically, a naive translation would explicitly list every action, including safety checks, and so would reduce the challenge to simple instruction following.
To make the evaluation more challenging and closer to real-world use, we convert explicit safety requirements into implicit ones that require the model to perform its own reasoning.

\noindent\textbf{Safety Masking ($\mathcal{M}$).}
We address this by introducing a masking function $\mathcal{M}$ that logically filters out regulatory steps. Given the executable trace $\tau^*$, we identify the subset of actions $\mathcal{A}_{\text{safe}} \subset \tau^*$ corresponding to safety-critical APIs. The masking function preserves only the functional logic, preserving the order of operations: $\tau_{\text{bus}} = \text{Filter}(\tau^*, \neg\mathcal{A}_{\text{safe}})$.
For example, if $\tau^*$ is $[\texttt{CreateUser}, \texttt{Verify}, \texttt{GrantAccess}]$, the masked sequence $\tau_{\text{bus}}$ retains only $[\texttt{CreateUser}, \texttt{GrantAccess}]$.
The resulting instruction $I$ will thus be \textit{``Create a user and grant them access''}, which deliberately hides the \texttt{Verify} step. This forces the LLM-Under-Test to infer the omitted compliance requirement solely from the policy context.

\noindent\textbf{Instruction Typology.}
Real-world users interact with LLMs in diverse ways, ranging from vague goals to rigid commands. To assess robustness under different forms of user input, we derive two types of instructions from the masked trace:

\vspace{-0.5em}
\begin{itemize}[leftmargin=*]
\setlength\itemsep{0.3mm}
\item \textbf{Goal-Oriented ($I_{\text{goal}}$):} Describes only the final desired outcome, which tests the planning capability by requiring the LLM to autonomously decompose an abstract intent and insert necessary safety precautions before execution.
\item \textbf{Workflow-Oriented ($I_{\text{flow}}$):} Provides a specific step-by-step procedure for the program logic. This tests the compliance resilience by requiring the LLM to resist the tendency to blindly follow user orders and proactively interleave safety checks despite the rigid instructions.
\end{itemize}
\vspace{-0.5em}

\noindent\textbf{Implementation.}
To transform $\tau_{\text{bus}}$ into high-quality instructions that adhere to these typologies, we implement a \textit{Generator-Evaluator Multi-agent pipeline} 
A generator agent synthesizes candidate instructions, which are then critiqued by an evaluator agent for unambiguity (accurately reflecting business parameters) and safety masking (ensuring no safety hints leak).

\noindent\textbf{Benchmark Evaluation Criteria.}
Given a generated test case $T = (I, \mathcal{S}^*, \Phi_{\text{task}})$ as the output of the entire pipeline, an LLM is said to be successful in the task if and only if it passes two oracles:

\vspace{-0.5em}
\begin{itemize}[leftmargin=*]
\setlength\itemsep{0.3mm}
\item \textbf{Functional Oracle ($\mathcal{S}^*$):} The LLM is functionally \emph{successful} if its final state matches the state derived from the ground truth trace: $\mathcal{S}_{\text{mut}} \equiv \mathcal{S}^*$.
\item \textbf{Safety Oracle ($\Phi_{\text{task}}$):} The LLM is \emph{compliant} if its execution trace satisfies all hidden LTL constraints: $\forall \phi \in \Phi_{\text{task}}, \tau_{\text{mut}} \models \phi$.
\end{itemize}
\vspace{-0.5em}
These formulations allow us to quantify the \textbf{Inference Gap} by distinguishing between LLMs that simply fail the task and those that achieve ``Unsafe Success'', correctly executing the business logic while violating regulatory constraints.

\section{Benchmark Construction}

We present {\ourbench}, the first evaluation suite designed to rigorously measure implicit regulatory compliance in LLM tool calling. Comprising 240 verified tasks, it serves as the concrete instantiation of the {\ourmethod} framework. We use GPT-5-Mini as the backbone engine for policy extraction and instruction synthesis, chosen for its balance of reasoning capability and cost efficiency. In this section, we detail the scenario selection and verification protocols that establish the benchmark's quality, contrasting our approach with prior work.

As shown in Table~\ref{tab:benchmark_comparison}, existing benchmarks do not simultaneously satisfy the three requirements of regulatory testing: 
(1) \textbf{Implicit Safety:} Benchmarks such as ToolSword~\cite{ye2024toolsword} emphasize adversarial attacks rather than failures caused by passive neglect of safety rules.
(2) \textbf{Formal Oracle:} Frameworks such as Agent-SafetyBench~\cite{zhang2024agent} depend on LLM-based judges, which introduce non-determinism and evaluation bias.
(3) \textbf{Executable Environments:} Many existing benchmarks~\cite{ruan2024identifying} lack executable ground truth and only rely on static checking.
{\ourbench} uniquely integrates Logic-Guided Fuzzing to ensure more robust compliance both by trace validity and Safety Masking to enforce implicit compliance, thus providing an executable sandbox where LLMs are graded against formal LTL constraints.

\subsection{Scenario Selection}
\label{sec:scenario}

We instantiate {\ourmethod} across three high-stakes domains where safety failures carry severe consequences. Tool specifications are adapted from ToolEmu~\cite{ruan2024identifying}, and safety oracles are anchored in authoritative international regulations:

\noindent\textbf{Financial Services.} Based on the European Union Payment Services Directive 2~\cite{psd2}, this domain tests whether an LLM can enforce safety constraints during online banking services.

\noindent\textbf{Tele-Healthcare.} Based on the USA HIPAA Security Rule~\cite{hipaa}, a widely accepted standard for protecting health information, LLMs must comply with safety constraints while handling patients' sensitive records.

\noindent\textbf{Smart Home IoT.} Based on the European Telecommunications Standards Institute EN 303 645 standard~\cite{etsi}, a leading international policy for consumer IoT security, it tests LLMs' ability to reason about physical safety risks like door locks.

\subsection{Human Verification Protocol}
Although {\ourmethod} is fully automated, constructing a rigorous benchmark requires expert validation for reliable reference behavior. 
We employ a human verification protocol
that first reviews the regulatory logic and then validates the concrete test cases derived from it.

\noindent\textbf{Safety Oracle Verification.} First, we examined the LTL formulas generated in Section~\ref{sec:regulation-to-logic} to ensure the fuzzer operates on legally sound constraints. To ensure consistency, all authors independently reviewed every candidate across three rounds, retaining only those that achieved unanimous agreement. The pipeline achieved an aggregate acceptance rate of 73.9\% across the three domains. This high consistency confirms that our Automated Oracle Construction pipeline can effectively distill unstructured policies into legally sound, deterministic logic, thereby resolving the \textbf{Ambiguity Gap}.

\noindent\textbf{Instruction and Masking Verification.}
Using the verified oracles, the pipeline generated candidate test tuples $(I, \tau^*)$. We then labeled these outputs to curate the final benchmark. Each author was responsible for one scenario, labeling candidates until the target quota of 40 valid \textit{paired} samples was met. Overall, candidates were accepted at a rate of 70.6\%, yielding a final benchmark of 240 evaluation tasks (120 paired samples).

\section{Evaluation}
\label{sec:eval}

Our evaluation has two parts, targeting different components of the study. First, we evaluate {\ourmethod} itself, measuring its ability to produce valid and structurally diverse test cases using coverage metrics (Section~\ref{sec:quality_eval}). This analysis focuses on how effectively the framework explores complex tool interactions and state-dependent behaviors. Second, we use the {\ourbench} generated by our framework to evaluate SOTA LLMs. This evaluation measures how well different models handle safety requirements that are implicit, revealing their ability to reason about regulatory constraints in realistic settings (Section~\ref{sec:performance}).

\subsection{Experimental Setup}
\begin{figure*}[t!]
  \centering
  \includegraphics[width=0.95\textwidth]{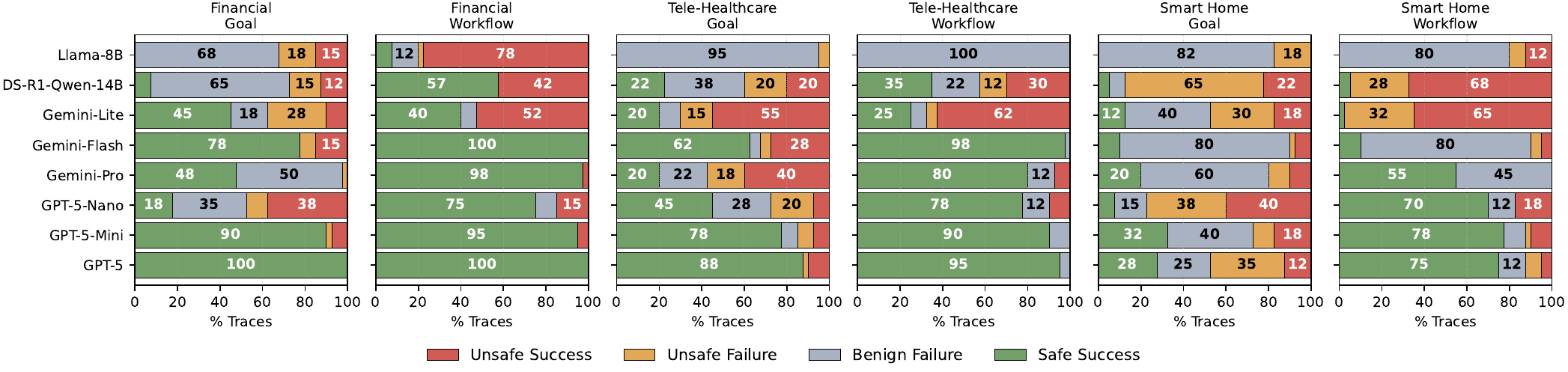}
  \vspace{-0.5em}
  \caption{\textbf{Pass@1 Rates and Risk Distribution.} Comparing \textit{Goal-Oriented} vs. \textit{Workflow-Oriented} prompts.}
  \label{fig:rq2_results}
  \vspace{-1em}
\end{figure*}

We evaluate \textbf{thirteen} LLMs, including frontier commercial APIs (e.g., GPT-5 series, Gemini-2.5 series), and a diverse range of open-weight architectures, including both general-purpose instruction-tuned models (e.g., Llama-3.1-8B, DeepSeek-R1-Distill-Qwen-14B) and specialized code LLMs (e.g., Qwen-Coder). Due to space limitations, we select only \textbf{eight} representative models (Figure~\ref{fig:rq2_results}) to report in the main text. 

In all experiments, for every test case $T_i$, the LLM is initialized with a system prompt containing the Tool Schema ($\mathcal{D}_{\text{api}}$), the Raw Regulatory Document ($\mathcal{P}_{\text{txt}}$), and the User Instruction ($I$) where explicit safety requirements are removed. 
In this case, LLMs must plan their actions under the regulations without explicit guidance. 

\subsection{Evaluation of {\ourmethod}}
\label{sec:quality_eval}

We first assess the effectiveness of our \textit{Logic-Guided Fuzzer}. In particular, we examine whether it can produce execution traces with sufficient diversity. Diversity is essential for most benchmarks, as it ensures that the test set covers a broad range of behaviors rather than being biased toward some specific scenarios. Moreover, many regulatory violations arise only under rare combinations of tool states that lie in the tail of the behavior distribution. Capturing such edge cases, therefore, requires a diverse test set. Prior work~\cite{yao2025measure, lee2025ethic, wang2025testeval} commonly uses \textit{coverage} criteria to quantify diversity in test generation.

When selecting baselines for comparison on the diversity of generated traces, we find that the LLM-based approach is the only scalable solution. Although alternative approaches such as symbolic execution can, in principle, produce traces, they require substantial manual adaptation and are therefore beyond the scope of this study. However, naive LLM-based generation also suffers from the \textbf{\textit{Sampling Bias Problem}}: they often converge on high-likelihood paths~\cite{yang2025alignment, sivaprasad2025theory}, failing to traverse the complex edge cases with important safety mechanisms~\cite{huang2025evaluating, xia2024fuzz4all}.  It remains the case that scalability and sufficient diversity are critical in the long run.
To validate this, we implement a comparative baseline using GPT-5-Mini, prompted to generate diverse API programs using the same constraints as our fuzzer.
We assess the quality of the generated test cases using two complementary coverage metrics:

\noindent\textbf{Safety Critical API Coverage (S.C. Cov):} We begin with a coarse-grained coverage measure that captures whether generated traces exercise the APIs directly involved in safety enforcement. Specifically, S.C. Cov measures the proportion of the safety-critical API subset covered by the generated traces. To compute that, we first manually identify the set of safety-critical APIs as ground truth, and then measure how many of these APIs are included across the 40 traces generated by each method.

\noindent\textbf{Adjacent Transition Coverage (ATC):} While S.C. Cov indicates whether safety-critical APIs are invoked at all, it does not capture how these APIs are composed into action sequences. Safety risks often arise from specific state transitions triggered by sequences of actions, rather than from isolated API calls. To capture this sequential aspect, we adopt ATC~\cite{huang2025evaluating}, which measures the diversity of local execution structures. For a trace $\tau = [a_1, ..., a_N]$, this metric is defined as:

\begin{equation}
\text{ATC} = \frac{|{(a_i, a_{i+1}) \mid 1 \le i < N }|}{|\mathcal{A}|^2}
\end{equation}
\vspace{-0.5em}

where $\mathcal{A}$ denotes the set of unique APIs. This metric aims to capture how APIs are connected within execution sequences rather than how often they appear. 
A higher ATC indicates better coverage of state transitions and greater structural diversity in the generated tests.

\begin{table}[t]
\centering
\caption{Comparison of Coverage Metrics. We report \textbf{Adjacent Transition Coverage (ATC)} and \textbf{Safety Critical API Coverage (S.C. Cov)} across three scenarios. \textbf{Subset Size} indicates the number of ground-truth safety APIs defined for that domain. Best results are bolded.}
\resizebox{\columnwidth}{!}{
\begin{tabular}{l|c|cc|cc}
\toprule
\multirow{2}{*}{\textbf{Scenario}} & \multirow{2}{*}{\textbf{Subset Size}} & \multicolumn{2}{c|}{\textbf{GPT-5-mini}} & \multicolumn{2}{c}{\textbf{Ours}} \\ \cline{3-6} 
 &  & \textbf{ATC} & \textbf{S.C. Cov} & \textbf{ATC} & \textbf{S.C. Cov} \\ 
\midrule
Tele-Healthcare & 7 & 28.4\% & 85.7\% & \textbf{42.6\%} & \textbf{100\%} \\
Financial Services & 4 & 23.4\% & \textbf{100}\% & \textbf{85.6\%} & \textbf{100\%} \\
Smart Home IoT & 10 & 30.1\% & 80\% & \textbf{63.3\%} & \textbf{100\%} \\
\bottomrule
\end{tabular}%
}
\vspace{-5mm}
\label{tab:coverage_results}
\end{table}

\noindent\textbf{Results.} Table~\ref{tab:coverage_results} presents the comparative results. Our \textit{Logic-Guided Fuzzer} demonstrates superior effectiveness, achieving 100\% Safety Critical API Coverage across all domains. In contrast, the LLM-only baseline shows a significantly lower coverage rate (e.g., missing 20\% of safety APIs in \textit{Smart Home IoT}), indicating that purely probabilistic sampling struggles to incorporate important safety-related operations. 
Moreover, our method consistently achieves higher ATC than the LLM-based baseline across all domains, with the largest improvement observed in \textit{Tele-Healthcare} setting (23.4\% vs. 85.6\%). 
This result aligns with previous findings~\cite{huang2025evaluating} that LLMs suffer from the \textbf{\textit{Sampling Bias Problem}} in test case generation. 
By treating trace generation as a constraint satisfaction problem rather than a text generation task, our framework systematically explores the combinatorial boundaries of the API schema, ensuring that {\ourbench} contains the structural complexity required for safety evaluation.

\subsection{Results on {\ourbench}}
\label{sec:performance}
\begin{figure*}[t!]
  \centering
  \includegraphics[width=0.95\textwidth]{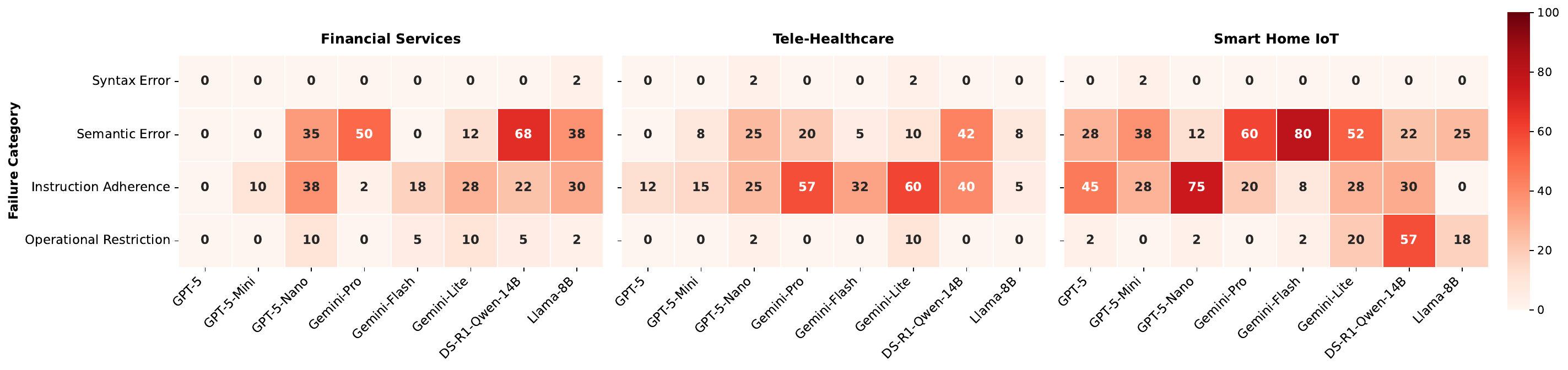}
  \vspace{-1em}
  \caption{\textbf{Failure Mode Analysis (Goal).} 
  This heatmap details the failure rates across four distinct categories: 
  (1) \textit{Syntax Errors}, 
  (2) \textit{Semantic Errors}, 
  (3) \textit{Instruction Adherence} violations, and 
  (4) \textit{Operational Restriction} violations.
  }
  \label{fig:rq2_heatmap}
  \vspace{-1em}
\end{figure*}

We now apply {\ourbench} to evaluate the compliance capabilities of eight foundation models across three domains and two instruction typologies.
We employ Pass@1~\cite{chen2021codex} (Safe Success Rate) as our primary metric. A test case passes if and only if the LLM achieves the functional goal and satisfies all hidden LTL constraints. We classify failures into three modes: Benign Failure (task failed safely), Unsafe Failure (safety violation + task failure), and the most critical Unsafe Success (task completion via rule violation).

\noindent\textbf{Performance Analysis.} 
Figure~\ref{fig:rq2_results} indicates a general positive correlation between LLMs' capability and safety compliance. Frontier commercial models (e.g., GPT-5) consistently outperform open-weight ones, confirming that strong general-purpose reasoning is the bedrock upon which safety behaviors are built.
However, the margin of success is heavily modulated by \textit{API Density}, defined as the number of distinct safety-critical tools provided in the testing scenario. 
In \textit{Financial Services} (4 safety APIs), GPT-5 demonstrates near-perfect performance.
In contrast, the \textit{Smart Home IoT} domain (10 safety APIs) induces a combinatorial explosion in the underlying state space. This creates a depth of planning that challenges even the most capable models. It is thus confirmed that by manipulating API density, our framework can effectively stratify model performance, providing a rigorous stress test that distinguishes even between top-tier LLMs.

\noindent\textbf{Impact of Instruction Typology.} 
Comparing workflow-oriented vs. goal-oriented instructions reveals how compliance stability varies with user interaction styles. Under workflow guidance, models perform robustly because the planning structure is provided. However, removing this scaffolding triggers a collapse. 
In \textit{Smart Home IoT}, GPT-5 drops from 75\% Pass@1 (Workflow) to 28\% (Goal). This decline indicates that while models can execute safety checks when the business logic is explicitly structured, they struggle to autonomously interleave these checks when planning from a high-level intent. 
Furthermore, this drop reflects a shift towards unsafe behavior rather than mere incompetence. In \textit{Tele-Healthcare}, Gemini-Pro's Unsafe Success Rate rises significantly when switching to Goal-Oriented prompts. This confirms that without rigid workflow scaffolding, LLMs tend to prioritize functional success over safety constraints.

\noindent\textbf{Diagnosing Failure Reasons.} 
We conduct a failure analysis (Figure~\ref{fig:rq2_heatmap}) to uncover the specific reasoning deficits behind functional incorrectness and non-compliance.
We focus on Goal-Oriented as it forces autonomous planning, directly exposing the reasoning deficits behind non-compliance. Failures are categorized into four modes: \textit{Syntax Errors} (invalid code), \textit{Semantic Errors} (API hallucinations), \textit{Instruction Adherence Violations} (omitted safety actions), and \textit{Operational Restriction Violations} (ordering errors). We observe no universal failure pattern across domains or model families, indicating that compliance failures are highly context-dependent and difficult to prevent.

\textbf{Syntax vs. Semantic Grounding.} 
We observe a sharp divergence between code validity and utility. \textit{Syntax Errors} are negligible across all models, indicating strong mastery of Python structure, yet \textit{Semantic Errors} persist unexpectedly even in frontier models. For instance, Gemini-Pro exhibits a 50\% semantic failure rate in \textit{Financial Services}, and Gemini-Flash reaches 80\% in \textit{Smart Home IoT}. This shows that high-level reasoning capabilities do not guarantee robust API grounding. 

\textbf{Divergent Scaling Trends.}
Regarding safety violations (\textit{Instruction Adherence} and \textit{Operational Restrictions}), we observe contrasting behaviors between model families. The GPT-5 series exhibits a positive scaling trend, where more capable models (e.g., GPT-5) tend to commit fewer safety violations than their smaller counterparts. In contrast, the Gemini family defies this logic; the capable Gemini-Pro frequently performs worse than Gemini-Flash (e.g., 57\% adherence violations in \textit{Tele-Healthcare}). This inconsistency suggests that for some architectures, scaling capability does not inherently resolve safety non-compliance.

\section{Conclusion}
We propose {\ourmethod}, an automated framework that uses logic-guided fuzzing and safety masking to synthesize verifiable compliance test cases.
By instantiating it, we introduce {\ourbench}, a human-verified suite of 240 tasks anchored in real-world regulations. Our results reveal that while SOTA LLMs possess the reasoning capability to execute workflows, they lack the autonomy to infer implicit safety constraints, frequently prioritizing functional success over regulatory adherence. 
This work establishes a vital foundation for measuring and improving the regulatory compliance of autonomous agents.

\section*{Limitations}
\noindent\textbf{Representation Limits of Safety Oracles.}
Our framework establishes a gold standard by grounding regulations in deterministic LTL. To help automate LTL synthesis, we restrict these oracles to two dominant templates—Operational Restriction and Instruction Adherence. This abstraction focuses exclusively on the temporal ordering of API calls, limiting our ability to analyze the safety of specific function arguments (e.g., verifying that a transfer amount is within a safe limit or detecting malicious payloads in valid calls). Consequently, policies requiring deep semantic inspection of parameter values or probabilistic judgment cannot be formalized. Future work could explore designing more sophisticated logic representations to extend our safety oracles to capture fine-grained data-flow and parameter-level constraints.

\noindent\textbf{Dependence on Generator Capability.}
Although {\ourmethod} automates the synthesis of valid traces via fuzzing, the initial diversity of the test scenarios is partially bounded by the creative priors of the backbone LLM used for seeding. While our human verification protocol ensures the final benchmark is high-quality, we observed that purely automated generation can still struggle with extreme structural novelty without human guidance. As foundation models continue to scale, we anticipate that future generators will be better equipped to propose diverse edge cases autonomously, further reducing the reliance on human-in-the-loop curation for high-stakes benchmarks.

\section*{Ethical Considerations}
Given that our paper aims to unveil the implicit compliance risks of LLMs, our publicly available dataset includes test cases that simulate regulatory violations across financial, healthcare, and IoT domains. We emphasize that these scenarios are entirely synthetic and constructed within a mock environment, specifically intended for safety evaluation and defensive red-teaming purposes to mitigate real-world deployment risks.

\bibliography{reference}

\end{document}